\documentclass[letterpaper, 10 pt, conference]{ieeeconf}  

\IEEEoverridecommandlockouts                              

\usepackage{graphicx} 
\usepackage{amsmath} 
\usepackage{amssymb}  
\usepackage{subcaption}

\usepackage{multirow}
\usepackage{adjustbox}
\newcommand{\fig}[1]{Fig.~\ref{fig:#1}}

\newcommand{\tab}[1]{Table~\ref{tab:#1}}

\newcommand{\sect}[1]{Section~\ref{sec:#1}}

\usepackage{xcolor}

\newcommand{\vect}[1]{{\boldsymbol{\mathbf{#1}}}} 

\DeclareMathOperator*{\argmin}{argmin}
\graphicspath{{images/}}

\title{\LARGE \bf
Photometric single-view dense 3D reconstruction in endoscopy
}

\author{V\'ictor M. Batlle, J.M.M. Montiel,~\IEEEmembership{Member,~IEEE} and Juan D. Tard\'os,~\IEEEmembership{Fellow,~IEEE}
\thanks{*~The authors are with the Instituto de Investigaci\'on en Ingenier\'ia de Arag\'on (I3A), Universidad de Zaragoza, 
Mar\'ia de Luna 1, 50018 Zaragoza, Spain. 
    E-mail: \{vmbatlle, josemari, tardos\}@unizar.es.}%
\thanks{This work was supported by EU-H2020 grant 863146: ENDOMAPPER, Spanish government grant PGC2018-096367-B-I00 and FPU scholarship of V. M. Batlle, and by Aragón government grant DGA\_T45-17R.}%
}

\begin{document}

\maketitle
\thispagestyle{empty}
\pagestyle{empty}

\begin{abstract}
Visual SLAM inside the human body will open the way to computer-assisted navigation in endoscopy. 
However, due to space limitations, medical endoscopes only provide monocular images, leading to systems lacking true scale. 
In this paper, we exploit the controlled lighting in colonoscopy to achieve the first in-vivo 3D reconstruction of the human colon using photometric stereo on a calibrated monocular endoscope. Our method works in a real medical environment, providing both a suitable in-place calibration procedure and a depth estimation technique adapted to the colon's tubular geometry.
We validate our method on simulated colonoscopies, obtaining a mean error of 7\% on depth estimation, which is below 3~mm on average. Our qualitative results on the EndoMapper dataset show that the method is able to correctly estimate the colon shape in real human colonoscopies, paving the ground for true-scale monocular SLAM in endoscopy.
\end{abstract}

\section{INTRODUCTION}

With the goal of improving the efficiency and effectiveness of routine diagnostic and medical intervention procedures, there is a growing research interest in extending augmented reality and autonomous navigation to the human body. These advances will need to accurately solve localization and mapping from visual sensors.

Simultaneous Localization and Mapping (SLAM) with stereo \cite{mur2017orb} and visual-inertial \cite{campos2021orb} cameras already provide great accuracy for multi-view reconstruction. In most medical endoscopy applications, space limitations restrict to monocular cameras. With monocular vision, the real scale of the environment cannot be observed, so potential applications are limited to up-to-scale reconstructions which usually suffer from scale drift problems, specially in deforming environments \cite{lamarca2020defslam}.

However, the interior of the human body is an example of an artificially illuminated environment, where the light source is controlled and linked to the camera movement. Our goal is to take advantage of the illumination to reconstruct 3D scenes, obtaining photometric stereo information by considering a light-camera pair.

\begin{figure}
    \centering
    \begin{subfigure}[b]{0.28\linewidth}
        \centering
        \includegraphics[width=0.9\linewidth]{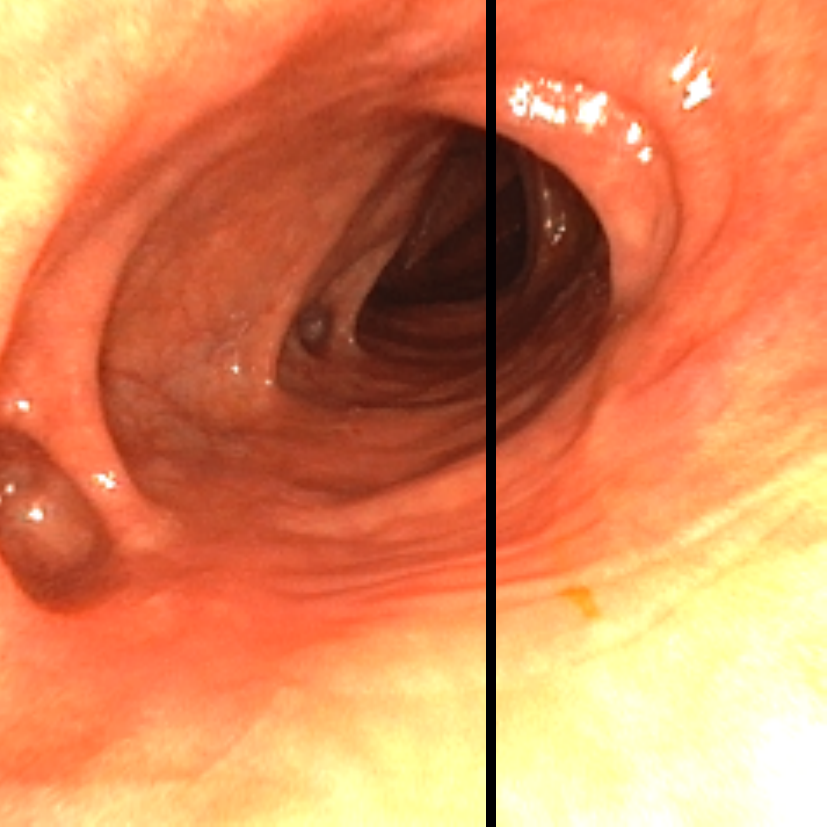}
        \caption{Input frame}
        \label{fig:hculb-results:original}
    \end{subfigure}
    \begin{subfigure}[b]{0.385\linewidth}
    \centering
        \centering
        \includegraphics[width=0.9\linewidth]{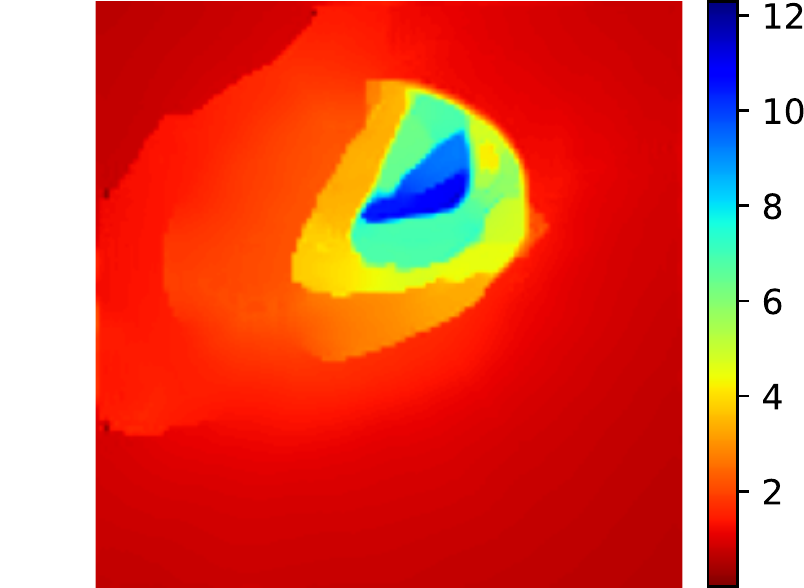}
        \caption{Est. depth [cm]}
        \label{fig:hculb-results:depth}
    \end{subfigure}
    \hfill
    \begin{subfigure}[b]{0.28\linewidth}
        \centering
        \includegraphics[width=0.9\linewidth]{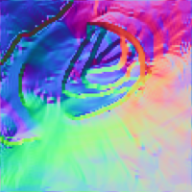}
        \caption{Est. normals}
        \label{fig:hculb-results:normals}
    \end{subfigure} \\
    \begin{subfigure}[b]{0.60\linewidth}
    \centering
        \centering
        \includegraphics[width=\linewidth]{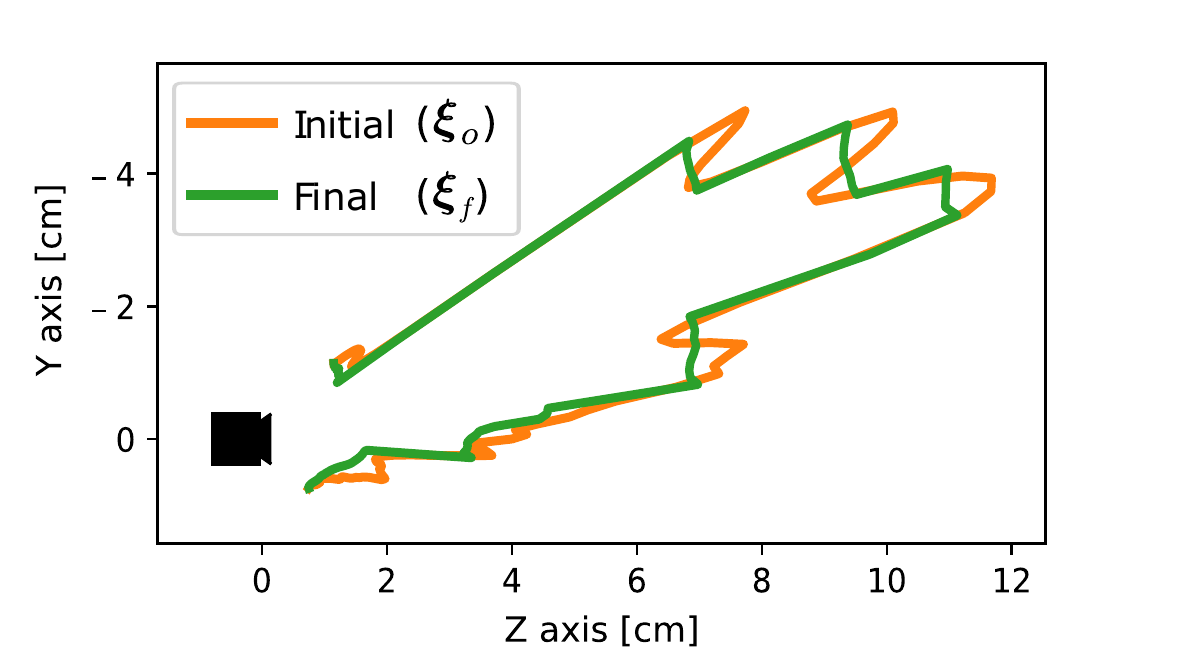}
        \caption{Cross-section along black line}
        \label{fig:hculb-results:section}
    \end{subfigure}
    \caption{Depth estimation on an 
    \emph{in-vivo} human colonoscopy.  
    }
    \label{fig:hculb-results}
\end{figure}

The main contributions of this paper are: (1) a simple photometric model for the endoscope light and camera, (2) a photometric calibration method that does not require a Lambertian pattern and can  be carried out in-place on a hospital setting, and (3) the first method capable of reconstructing the geometry of the human colon from a single view, only from the illumination of a calibrated conventional monocular endoscope (see~\fig{hculb-results}). This would solve the scale-drift problem in monocular SLAM, and can also provide real-scale maps when the albedo of the surface and the endoscope's auto-gain are known.

\section{RELATED WORK}

Recent results in single-view depth estimation using deep convolutional networks \cite{godard2019digging} open the possibility of designing accurate SLAM systems from monocular cameras, specially hybrid approaches \cite{yang2020d3vo} which combine deep learning with traditional methods. Previous work demonstrates that, using a depth estimation network, it is possible to perform scale-aware monocular SLAM, obtaining almost the same accuracy as with stereo, and eliminating scale drift \cite{li2018scale,tiwari2020pseudo,martinez2020scale,campos2021scaleaware}. 

A first attempt to apply these methods to endoscopy sequences achieves real-scale reconstructions with good accuracy \cite{recasens2021endo}. However, these methods require stereo supervision to learn how to predict the true size. Thus, today its application in colonoscopy is limited by the impossibility of acquiring stereo images of the human colon.

Other authors focus on the study of photometry to obtain dense and semi-dense reconstructions of outdoor and indoor 3D scenes \cite{newcombe2011dtam, engel2017direct}. They assume constant illumination, usually ambient light, ignoring any change in lighting.

In contrast, inside the human body, the illumination is controlled and light moves together with the camera. Recent work \cite{modrzejewski2020light, hao2020photometric} shows that changes in lighting, instead of being ignored, can be used to our benefit, obtaining dense reconstructions from monocular sequences.

\subsection{Lighting model}

Previous works propose a lighting model for their working environment. Specifically, they model light emission, interaction with surfaces, and capture by the camera. So far, the complexity of these \emph{ad-hoc} models required them to be calibrated and tested in laboratory environments. In this paper, we propose a simplified model that allows easier calibration without the need for Lambertian patterns.

{Modrzejewski et al.}~\cite{modrzejewski2020light} do a thorough work on analyzing various light source models. Their Spot Light Source (SLS) model offers a good compromise between complexity and accuracy. We adopt a similar approach, but with the aim of modeling the multiple light sources of the endoscope as a single virtual light. This leads us to a generic model in which our virtual light is located at the camera's optical center.

{Hao et al.}~\cite{hao2020light} calibrate the light emission separately by means of a plane mirror. Conversely, we propose a joint calibration method, which also allows easy estimation of camera geometry and photometry at the same time. 


A common approach \cite{modrzejewski2020light, visentini2015simultaneous} consists of assuming Lambertian surfaces, both during calibration and during reconstruction inside the human body. However, we show that this causes bias in the calibration when the real surface is not perfectly Lambertian. If not corrected, this error propagates to the 3D reconstruction. In contrast, our calibration considers non-Lambertian properties, giving results that are not affected by the calibration pattern used.

\subsection{Reconstruction}

In photometric stereo, the discussed light model can be used to obtain a dense depth map of the scene. The main discrepancy between the different approaches lies in their corresponding reconstruction method. 

{Modrzejewski et al.} \cite{modrzejewski2020light} propose an initial multi-view reconstruction followed by a photometric optimization, where a regularization term tends to favor smooth planar surfaces. Crucially, the multi-view method requires a rigid environment. In contrast, our method is based only on lighting, being able to reconstruct the environment from a single view. In addition, the geometry of the human colon, with numerous discontinuities, is far from planar. By considering this in our regularization, we can reconstruct its complex shape.

{Hao et al.} \cite{hao2020photometric} focus on specular highlights, where their method achieves the best accuracy. However, the accuracy of their reconstruction decreases for the rest of the continuous surface. Unlike them, we perform a global optimization, considering each point equally, which does not require the surface to be continuous.

To the best of our knowledge, previous work on dense photometric reconstruction on endoscopy \cite{modrzejewski2020light, hao2020photometric, okatani1997shape, collins2012towards} has been validated on nearly planar scenes, without discontinuities. Focusing on colonoscopy, Parot et al. \cite{parot2013photometric} provide experimental validation on phantoms. Instead, we demonstrate that our method can recover for the first time the tubular topology of a human colon, form a single {\em in-vivo} video frame, preserving the anatomical folds of the intestine, known as haustra.

\section{ENDOSCOPE MODEL}

\begin{figure*}
    \centering
    \begin{tabular}{ccc}
        \includegraphics[width=0.44\linewidth]{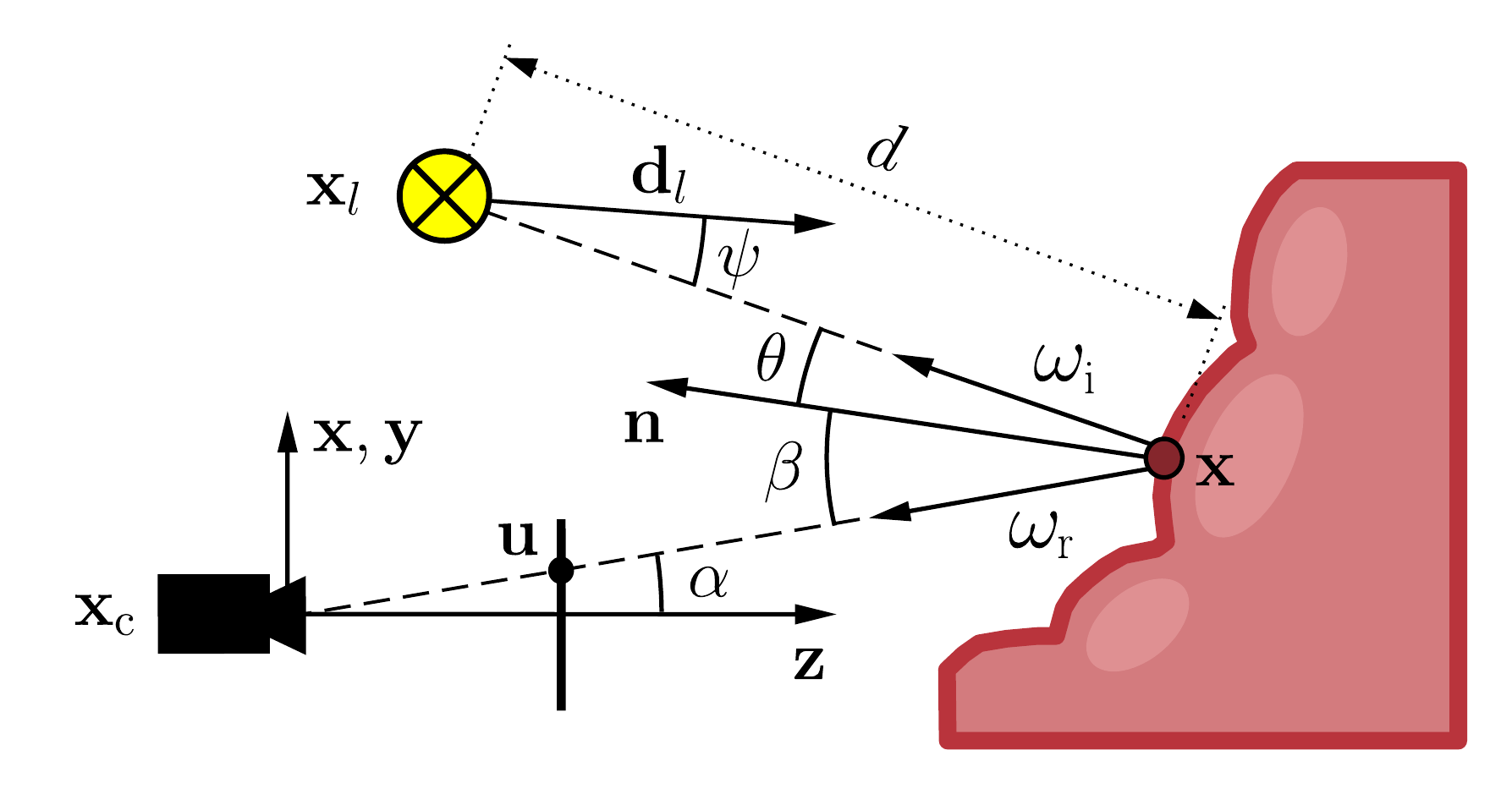} & \hspace{0.5cm} &
        \includegraphics[width=0.44\linewidth]{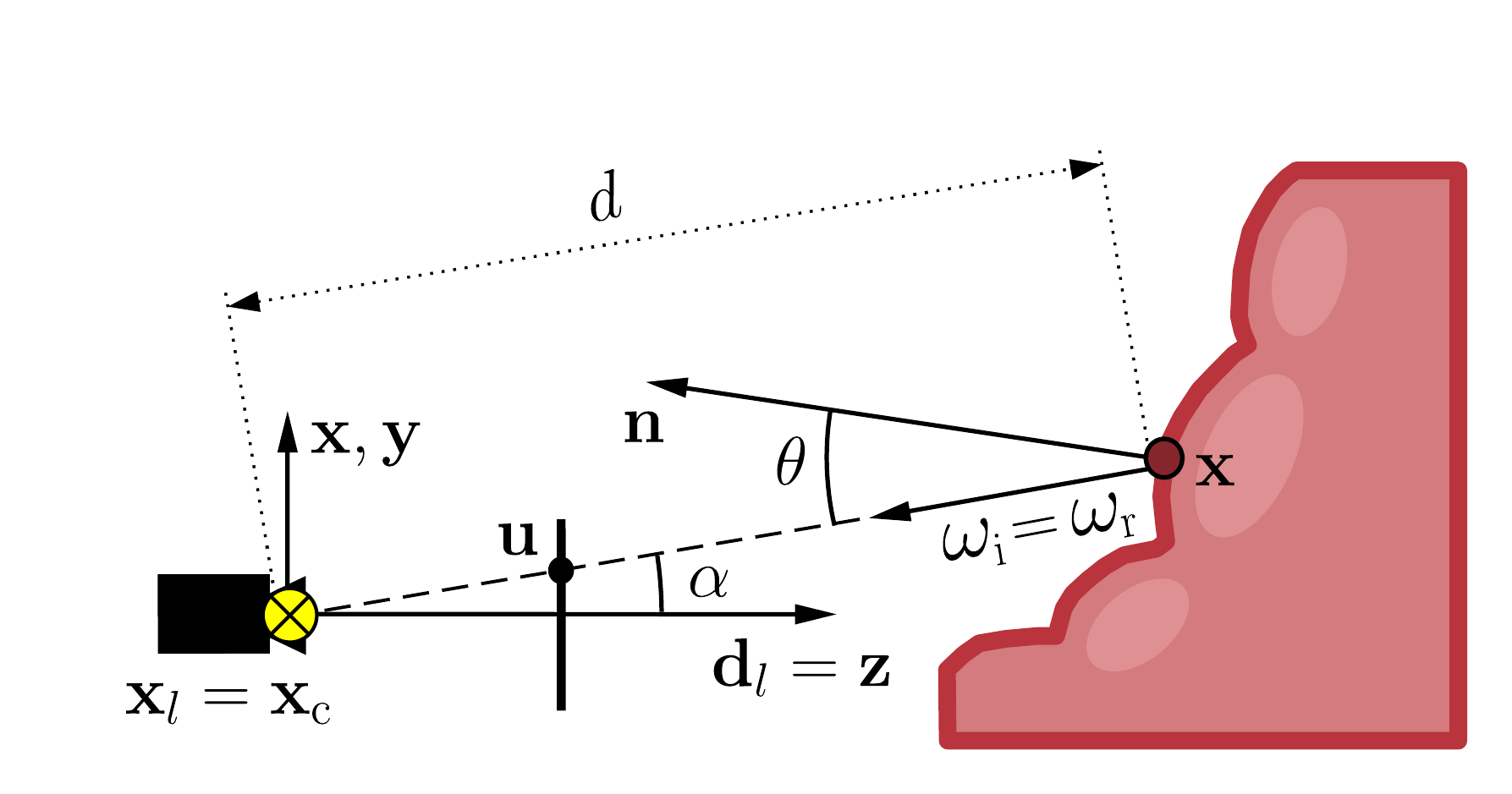}
    \end{tabular}
    \caption{\textbf{Left:} General photometric model. \textbf{Right:} Simplified photometric model. We assume a virtual light is located at the camera optical center and the light's principal direction is at the camera forward vector.}
    \label{fig:symbols-overview}
\end{figure*}

This paper presents a photometric approach to the problem of monocular 3D reconstruction during medical endoscopy. This approach considers the geometric model of image formation and the photometric model of light transport (\fig{symbols-overview}).

\subsection{Geometric model}
An endoscope camera is designed to cover a wide view angle. Thus, we adopt the approach of Kannala \& Brandt \cite{kannala2006generic} to model the fisheye lens, with four projective parameters $f_x, f_y, C_x, C_y$, and four distortion coefficients $k_{1-4}$. We denote the optical center of the camera as $\mathbf{x}_\text{c}$.

\subsection{General photometric model}
The illumination system mounted on an endoscope usually consists of one or more small lights. Given its small size, these lights are usually modeled as punctual lights \cite{modrzejewski2020light, hao2020photometric}. Thus, radiance $L_\text{i}$ coming from light center $\mathbf{x}_l$ to a point $\mathbf{x}$ in a surface, is subject to the inverse-square law:
\begin{equation}
    L_\text{i}(\mathbf{x}) = 
    \mu(\mathbf{x}) \frac{\sigma_o}{\lVert \mathbf{x} - \mathbf{x}_l \rVert ^ 2}
\end{equation}

In most endoscopes light is transmitted to the tip using optical fiber. Therefore, the amount of light emitted in each direction of space is not uniform. We model this behavior with a light spread function $\mu(\mathbf{x})$, that can be specified by fixing a principal direction $\mathbf{d}_l$, over which maximum radiance $\sigma_\text{o}$ is emitted, and assuming a radial cosine fall-off \cite{hao2020photometric}. We decide to modulate this decay by adding the cosine exponent $k$ as a parameter:
\begin{equation}
    \mu(\mathbf{x}) =
    \cos^k \psi, \quad 
    \psi = \langle \mathbf{x} - \mathbf{x}_l, \; \mathbf{d}_l \rangle
\end{equation}

When light reaches a surface, most of it will be reflected, going out in different directions depending on the material properties. The Bidirectional Reflectance Distribution Function (BRDF)  $f_{\text{r}}(\omega_{\text{i}},\omega_{\text{r}})$ defines how light is reflected at an opaque surface. Usually, this behavior depends on the incoming $\omega_\text{i}$ and outgoing $\omega_\text{r}$ direction of the light ray with respect to the normal $\mathbf{n}$ of the surface at that point. The inclination of the incident ray modifies the area of the projection of the solid angle on the surface, depending on the cosine of its angle with the normal. As a result, the reflected radiance is:
\begin{equation}
    L_\text{r}(\mathbf{x}, \omega_\text{r}) = 
    L_\text{i}(\mathbf{x}, \omega_\text{i}) \,
        f_{\text{r}}(\omega_{\text{i}}, \omega_{\text{r}})
        \cos \theta
\end{equation}
where $\theta = \langle \omega_\text{i}, \mathbf{n} \rangle$ and, for our case, $\omega_\text{i}$ is the direction to the light source and $\omega_\text{r}$ points to the camera (see \fig{symbols-overview}).

Light reaching the camera is affected by a set of factors. The capture system usually introduces attenuation on the received radiance. Natural  vignetting  tends  to  approximate  to  $\cos^4 \alpha$, where $\alpha$ is the off-axis angle between the ray direction and the camera forward $\mathbf{z}$ \cite{szeliski2010computer}. Mechanical vignetting is not easy to model theoretically. Therefore, vignetting is usually empirically approximated \cite{engel2016photometrically}. We assume radial attenuation from the camera’s forward vector, by modeling the decay with a $k'$ exponent on a cosine function:
\begin{equation}
    V(\mathbf{x}) = \cos^{k'} \alpha, \quad 
    \alpha = \langle \mathbf{x} - \mathbf{x}_\text{c}, \; \mathbf{z} \rangle
\end{equation}

Endoscope cameras might automatically adjust some parameters, such as exposure time or signal  amplification. In  video  streams,  this is controlled by an automatic gain control (AGC) logic. We assume this auto-gain usually acts as a multiplying factor $g_t$ at each $t$-th time instant. In order to increase the perceived dynamic range, cameras map the captured values through a gamma function with a common value of $\gamma = 2.2$ that does not change over time.

Our complete photometric model considers all concepts introduced above, as a combination of light, surface and camera effects:
\begin{equation}
    \mathcal{I}(\mathbf{x}) = 
    \left( 
        \frac{\mu(\mathbf{x}) \; \sigma_\text{o}}
            {\lVert \mathbf{x} - \mathbf{x}_l \rVert ^ 2} \;
        f_\text{r}(\omega_\text{i}, \omega_\text{r}) \;
        \cos\theta \;
        V(\mathbf{x}) \;
        g_\text{t}
    \right) ^ {1/\gamma}
\end{equation}

\begin{figure*}
    \centering
    \includegraphics[width=\textwidth]{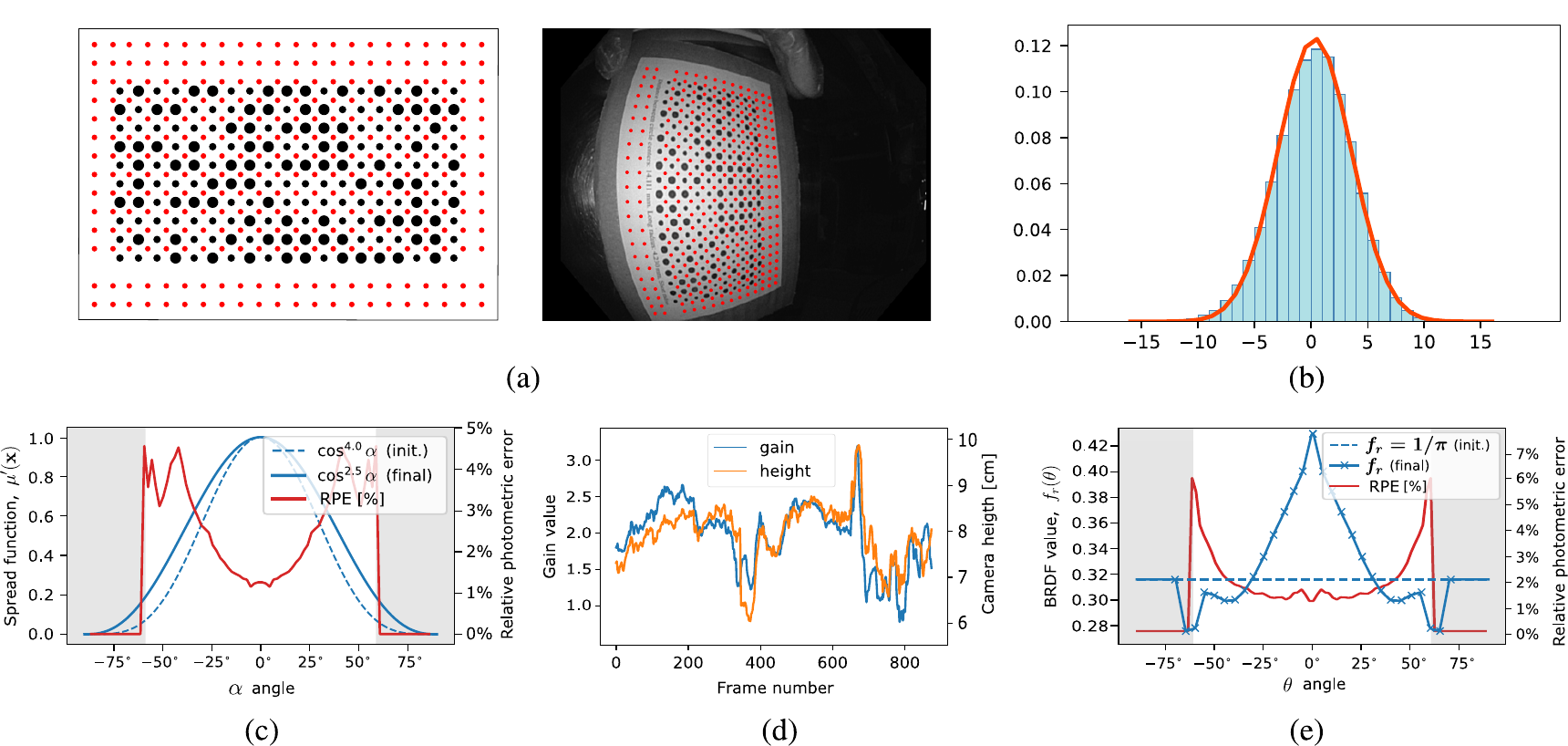}
    \caption{Sampling the Vicalib pattern: (a) Red marks correspond to each $\boldsymbol{x}_j$ sampled point. Photometric calibration results: (b) Photometric errors of the calibrated model are close to a Gaussian distribution with a mean of 0.3 and std. of 3.2 gray levels. (c) Joint attenuation caused by light spread function $\mu(\vect{x})$ and camera vignetting $V(\vect{x})$. (d) Estimated auto-gain factors over the calibration video. (e) Non-Lambertian BRDF for the paper sheet used for calibration.}
    \label{fig:calib-results}
\end{figure*}

\subsection{Simplified photometric model}

The presented model takes into account only one light source. Each additional light source must be modeled independently, in a similar way, but adding complexity to the model. Moreover, the characteristics of each endoscope version vary slightly. Commonly we find two or three optical fiber guides, which conduct the light to different points on the tip of the endoscope.

Instead of the costly process of modeling the details of each specific hardware, we propose a simplification based on encapsulating the joint effects of all the light points into a single virtual light source (see \fig{symbols-overview}). 
We observe that these light points are usually distributed fairly symmetrically around the endoscope camera. Therefore, we decided to place the virtual light source at the optical center of the camera, i.e. $\mathbf{x}_l = \mathbf{x}_\text{c}$ and we align light's principal direction with the camera forward, i.e. $\mathbf{d}_l = \mathbf{z}$.

In this new set-up, camera’s vignetting and virtual light's spread function are coupled, i.e. $\psi = \alpha$.  Thus, we model the effect of both functions jointly, as:
\begin{equation}
    \mu'(\mathbf{x}) = \mu(\mathbf{x}) \, V(\mathbf{x}) = \cos^k \alpha
\end{equation}

Regarding surface reflectance, now incoming $\omega_\text{i}$ and reflected $\omega_\text{r}$ directions match. Thus, the domain of the BRDF can be simplified. We will consider the incident angle $\theta$ of the light on the surface, such that the BRDF is simply $f_\text{r}(\theta)$.

In most endoscopes auto-gain logic is unknown. Therefore, $g_\text{t}$ values are coupled with the absolute radiance $\sigma_\text{o}$, so that their effects cannot be separated. Consequently, we fix the $\sigma_\text{o}$ parameter to an arbitrary value and estimate the relative auto-gain changes.

Finally, we obtain a simplified photometric model, which is parameterized according to the unknowns we want to estimate for our endoscope:
\begin{equation}
    \label{eq:simple-model}
    \mathcal{I} \left(
        \mathbf{x}, \,  k, \, g_\text{t}, \, \gamma
    \right) = 
    \left(
        \frac{\mu'(\mathbf{x}, \, k)}{\lVert \mathbf{x} - \mathbf{x}_l \rVert^2} \  
        f_\text{r}(\theta) \  
        \cos \theta \  
        g_\text{t}
    \right)^{1/\gamma}
\end{equation}

\section{ENDOSCOPE CALIBRATION}
\label{sec:calibration}

Geometric and photometric calibration is performed with a small Vicalib \cite{robotics2016perception} pattern printed on a white paper sheet of $5.61 \times 9.82$~cm. From a video captured with the endoscope, geometric parameters are obtained by processing 1 out of 20 frames using the Vicalib software.

Focusing on photometry, we propose an optimization problem that aims to minimize the photometric error between the empirical data $I$ and our model. For that, we select a set of $j$ sample points, uniformly distributed along the white areas of the pattern  (see \fig{calib-results}a). Then, the photometric loss function is computed on every visible $\boldsymbol{x}_\text{j}$ point on each $t\text{-th}$ frame of the video, using Huber function $\rho$ to be robust against spurious:

\begin{equation}
    \{k, g_\text{t}, \gamma \mid \forall \text{t}\}^* = 
    \argmin_{k, \, g_\text{t}, \, \gamma}
    \sum_{\text{j}, \text{t}}
    \rho \left( 
        I_{\text{jt}} - 
        \mathcal{I} \left(\boldsymbol{x}_\text{j}, k, g_\text{t}, \gamma \right)
    \right)
\end{equation}

\subsection{Results}

From a video resolution of $1440 \times 1080$~px at 30 frames per second, the calibration obtains values for the geometric parameters $f_x = 717.21$~px, $f_y = 717.48$~px, $C_x = 735.37$~px, $C_y = 552.80$~px and the four distortion coefficients $k_{1-4}$ are [$-0.13893$, $-1.2396E-03$, $9.1258E-04$, $-4.0716E-05$]. These lead to a reprojection error in RMSE of 0.5288~px.

Regarding the photometric calibration results, the optimization converges to $k = 2.5$, $\gamma = 2.2$ and estimates auto-gain $g_\text{t}$ values ranging from 1 to 3. We observe that the final spread is wider than the natural vignetting $\cos^4 \alpha$  (see~\fig{calib-results}c). This is consistent with the illumination system of our endoscope, where three real light sources result in a widening of our virtual light's spread. Validation results show a small relative error of 1\% in the center, i.e. $0^{\circ}$. However, error grows towards the edges, reaching 4.5\% at $40^{\circ}$ and when $\alpha > 60^{\circ}$ (shaded area) the function remains unsampled in our calibration data.

Automatic gain cannot be evaluated with test data, because each image has a different gain factor. Instead, we can see that the estimated gain factor follows a continuous progression over time along the calibration sequence  (see~\fig{calib-results}d). Moreover, the gain value for each frame seems to be closely related to the distance from the camera to the illuminated surface. That is, when the camera is closer to the pattern, light is more intense, and the endoscope applies a lower gain value.

In addition, we observed that modeling the paper reflectance with a Lambertian BRDF $f_\text{r}(\theta) = 1/\pi$ led to a biased calibration. Therefore, we decided to also optimize the value of the BRDF for fifteen values of the $\theta$ angle and apply linear interpolation in the rest of the domain of the function (see~\fig{calib-results}d).
The estimated BRDF for the paper sheet shows specular behavior when the camera is close to the perpendicular to the surface. This results in a peak in the reflectance when the $\theta$ angle is near zero.

The result of the calibration allows us to estimate the gray level of a pixel with a standard deviation of 3.2 levels. The distribution of errors is unbiased  (see~\fig{calib-results}b). Moreover, the new estimated BRDF is an isolated component of the model. Therefore, when we want to apply our calibration in the interior of the human body, we can replace this BRDF with that of the human colon, and the rest of the calibrated parameters remain valid.

\section{DEPTH ESTIMATION}

Given the calibrated endoscope photometric model and a single endoscope image, our goal is to estimate depth and surface normal for each imaged 3D point. We consider the following assumptions: 
\begin{itemize}
    \item Similarly to Modrzejewski et al. \cite{modrzejewski2020light}, we assume that human tissue can be approximated by a Lambertian material if specular highlights are masked or treated as spurious. For this, we propose an automatic method for highlight detection and inpainting.
    \item In addition, given the weak texture of the colon tissue, the surface albedo $k_d$ is measurable and considerably constant. In our experiments we set $f_\text{r}(\theta) = k_d/\pi$.
    \item The imaged surfaces are smooth, except at occasional discontinuities. This allows us to approximate differential changes of the surface by a tangent plane.
\end{itemize}

Based on DTAM method \cite{newcombe2011dtam}, we approach the estimation of a depth map as an optimization problem, that minimizes an energy function:
\begin{equation}
    E_{\boldsymbol{\xi}} =
    \int_{\Omega} \Big\{ 
        C(\mathbf{u}, \boldsymbol{\xi}(\mathbf{u})) + 
        \lambda R(\mathbf{u}, \boldsymbol{\xi}))
    \Big\} d \mathbf{u}
\end{equation}
\noindent where
\begin{itemize}
    \item $\mathbf{u} \in \Omega$ are coordinates on the image,
    \item $\boldsymbol{\xi}: \Omega \rightarrow \mathbb{R}$ is the depth map,
    \item $C(\,)$ is a photometric cost function,
    \item $R(\,)$ is a regularization cost,
    \item $\lambda \in \mathbb{R}^{+}$ adjusts the regularization weight.
\end{itemize}

\subsection{Photometric cost function}
\label{sec:photometric-cost}

DTAM assumes ambient light on the scene and uses a cost function based solely on camera geometry and brightness constancy. However, the illumination during endoscopy varies with camera movement. Consequently, we replace the original cost function with a novel cost function based on our photometric endoscope model:
\begin{equation}
    C(\mathbf{u}, d) = 
    \rho \left(
        I(\mathbf{u}) -
        \mathcal{I} \left(\pi^{-1} \left(\mathbf{u}, d\right) \right)
    \right)
\end{equation}
\noindent where
\begin{itemize}
    \item $d$ is the Euclidean distance to the world point,
    \item $\pi^{-1}(\,)$ is the camera unprojection model,
    \item $\mathcal{I}(\,)$ is our calibrated endoscope photometric model \eqref{eq:simple-model},
    \item $I: \Omega \rightarrow \mathbb{R}^{+}$ denotes the actual pixel intensity,
    \item $\rho(\,)$ is the Huber robust cost function.
\end{itemize}

\subsection{Normal estimation from a depth map}

The photometric model of a scene $\mathcal{I}$ is influenced by both the distance to the points (inverse-square law) and the surface normal (cosine term). However, surface normal is directly related to depth variations. Therefore, both parameters should not be optimized separately. Instead, given the local planarity assumption, we can calculate the normal of a point from the estimated depth map \cite{hinterstoisser2011multimodal}.

Thanks to this relationship, we keep the depth map as the only unknown variable of the problem. However, it should be noted that this method is influenced by spurious data, especially at surface discontinuities. Therefore, in these areas, we expect some localized errors.

\subsection{Smoothness regularization}

The defined cost function is trying to find three unknowns per pixel ($d, n_\theta, n_\varphi$) from one intensity measurement ($I$). In order to solve the problem's ill-posedness, DTAM proposes a regularization term that penalizes local depth variations, except at points where the luminosity gradient is large, which usually correspond to surface discontinuities:
\begin{equation}
    R(\mathbf{u}, \boldsymbol{\xi}) = g(\mathbf{u}) \lVert \nabla \boldsymbol{\xi} (\mathbf{u}) \rVert_{\epsilon}
\end{equation}
 Thus, $\lVert x \rVert_\epsilon$ Huber norm with $\epsilon \approx 1.0e^{-4}$ works as total variation (TV) regularizer and $g(\mathbf{u})$ reduces the regularization strength at high gradient points. Thanks to these two terms, $(d, n_\theta, n_\varphi)$ are now constrained by the pixel's neighborhood, and at the same time the discontinuities of the colon can be preserved.

However, this might not be the best regularizer in the colon's tubular geometry. The first derivative $\nabla\vect{\xi}$ always favors zero changes along the depth map. So the reconstruction will tend to a plane parallel to the camera. Instead, similarly to \cite{modrzejewski2020light}, we use the second-order derivative $\nabla^2\vect{\xi}$ to impose smoothness, although we continue to allow discontinuities.

\subsection{Depth map representation}

DTAM formulates its depth map $\boldsymbol{\xi}_{1/z}$ as the inverse distance in the Z-axis. This decision is appropriate for a multi-view-based problem, as the pinhole projection model depends directly on this variable. Instead, we are faced with a single-view problem. In our case, the photometry is quadratically dependent on the inverse of the Euclidean distance, i.e. $\mathcal{I} \propto {1}/{d^2}$.

Therefore, we will compare the previous formulation with two new variants of the depth map, such that
\begin{equation}
    \boldsymbol{\xi}_{1/z} = \frac{1}{z}, \qquad \boldsymbol{\xi}_d = d, \qquad \boldsymbol{\xi}_{1/d} = \frac{1}{d}
\end{equation}

\subsection{Initial solution}

We make the optimization method start from an initial solution, where we assume all surface normal vectors pointing towards the camera optical center.

From the calibrated photometric model, we revert the effects of light spread function, Lambertian BRDF, as well as known camera gain and gamma correction:
\begin{equation}
    I_\text{c}(\mathbf{u}) =
    \frac{I(\mathbf{u})^\gamma}{\mu'(\mathbf{x}) \cdot f_\text{r}(\theta) \cdot g_\text{t}} =
    \frac{\cos\theta}{d^2}
\end{equation}
where $I_\text{c}$ is a \emph{canonical intensity value}, which is obtained after compensating all mentioned parameters that influence image formation. Note that, when a surface normal points towards the camera, the $\theta$ angle is zero. Therefore, by solving for $d$ in the above equation, we get an initial solution
\begin{equation}
    d_o(\mathbf{u}) = I_\text{c}(\mathbf{u})^{-1/2}
\end{equation}

The closer the actual $\theta$ is to zero, the closer this initial solution is to the real depth.

\section{EXPERIMENTAL RESULTS}

In this section, we first conduct some simple experiments to determine the best smoothness regularizer and depth map variant. Then, we check the accuracy of our depth estimation method with a photo-realistic simulation of a human colon. Finally, we test our method in a real in-vivo colonoscopy image.

\subsection{Simple geometry dataset}
\label{sec:simple-dataset}


\begin{figure}
    \centering
    \includegraphics[width=0.9\linewidth]{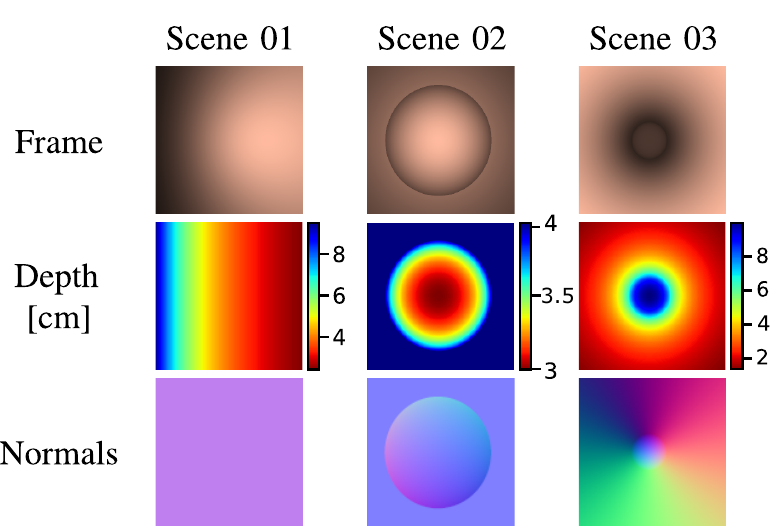}
    \caption{Data generated for our simple geometry dataset. \textbf{Top:} Frame simulated with our model as in~\eqref{eq:simple-model}. \textbf{Middle:} Ground-truth Z-depth map. \textbf{Bottom:} Ground-truth normal map, represented in color space $(R, G, B) = ([n_x, n_y, n_z] + 1) / 2$.}
    \label{fig:simple-dataset}
\end{figure}

\begin{figure}
    \centering
    \setlength\tabcolsep{0pt}
    \begin{tabular}{c c c}
        \\
        Scene 01 & Scene 02 & Scene 03 \\
        \includegraphics[height=2.35cm]{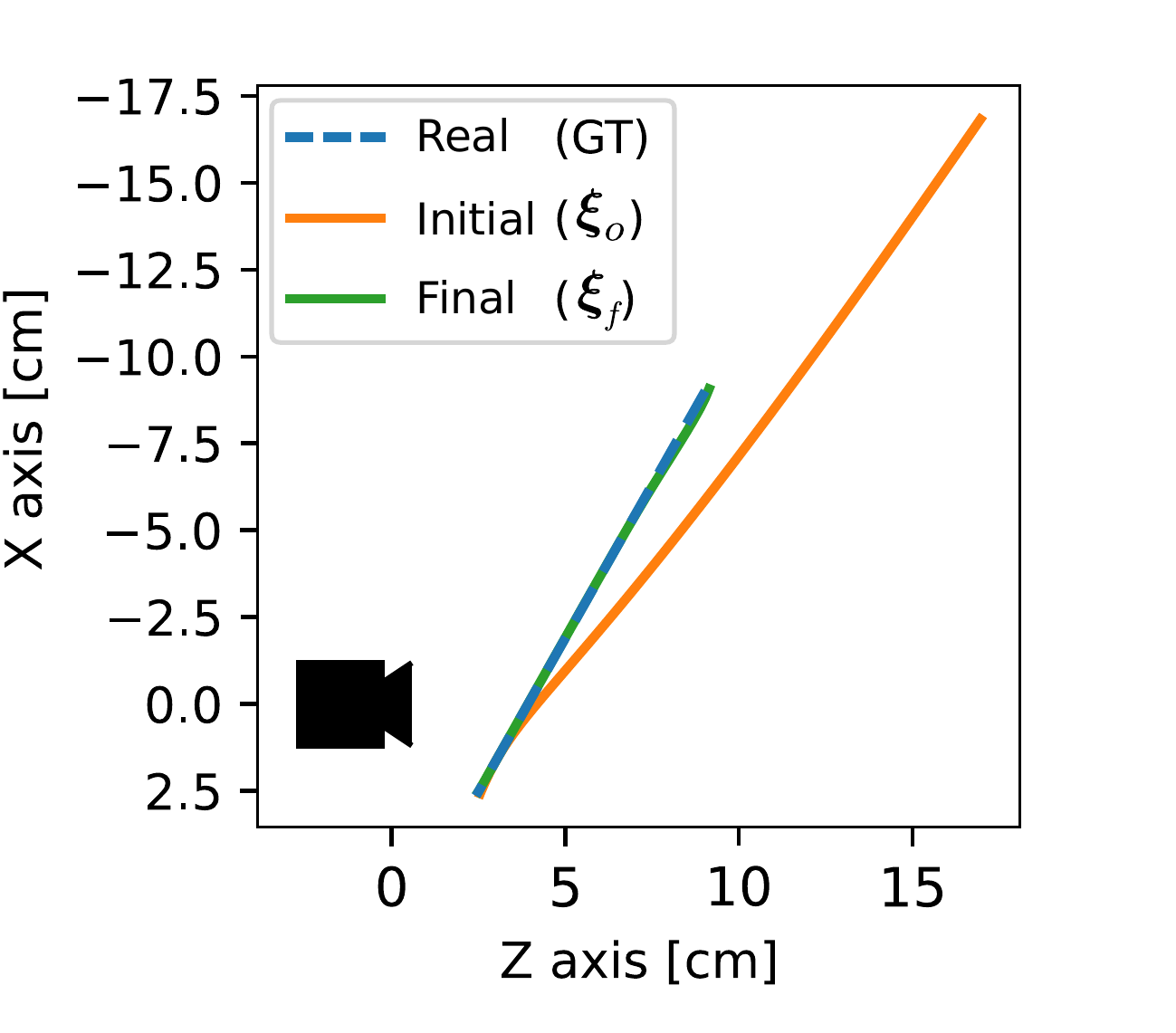} &
        \includegraphics[height=2.35cm]{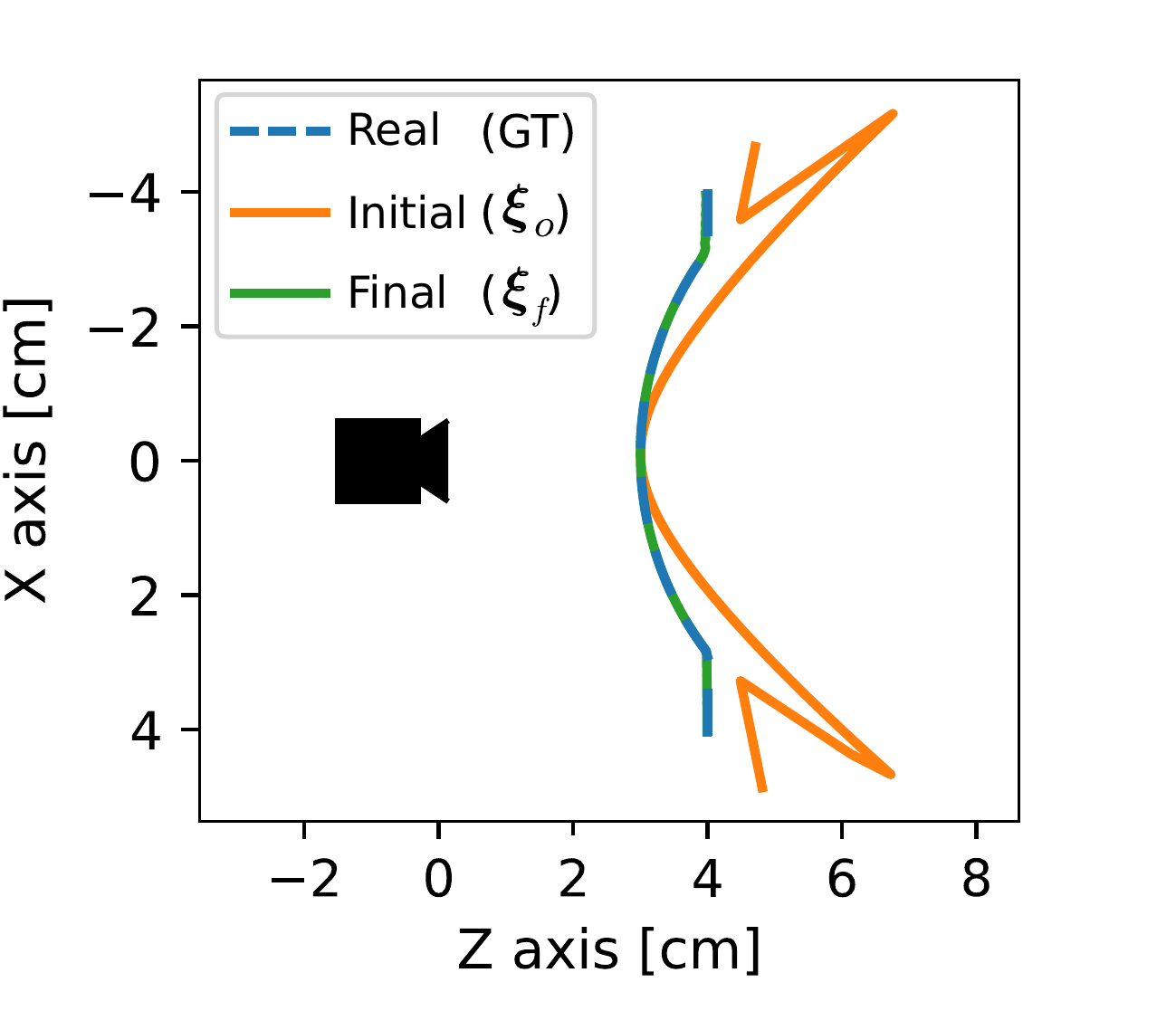} &
        \includegraphics[height=2.35cm]{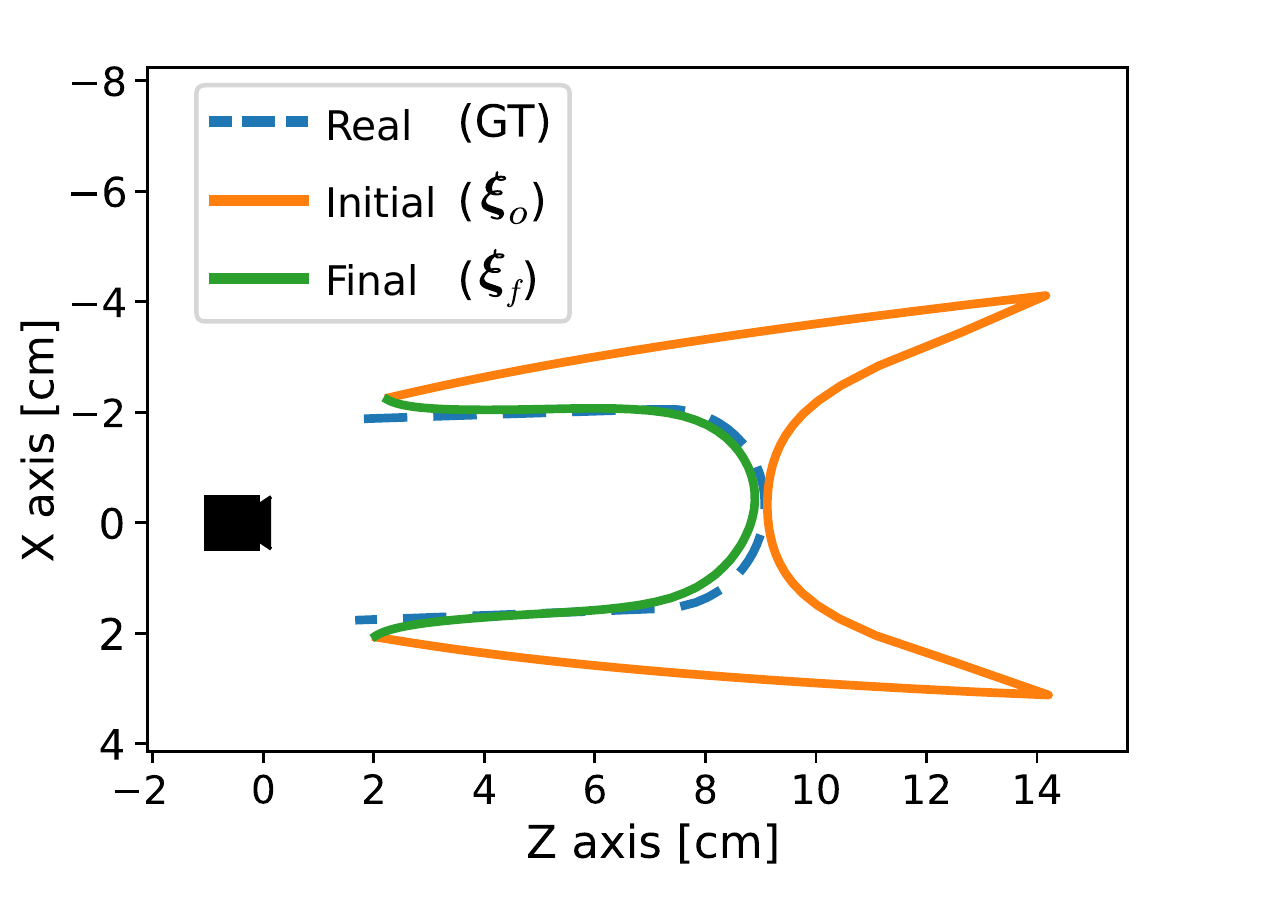}
    \end{tabular}
    \caption{Cross-section along Y-axis (middle of the image) of actual surface (GT), initial solution ($\vect{\xi}_o$) and final optimized depth map ($\vect{\xi}_f$). Our estimation matches the ground-truth.}
    \label{fig:simple-0to2-convergence}
\end{figure}


\begin{table}
\caption{\textsc{Accuracy on simple geometry dataset}}
\label{tab:simple-dataset}
\centering
\resizebox{\linewidth}{!}{%
\begin{tabular}{|c|c|c|r|rr|rr|rr|}
\hline
\multirow{2}{*}{\textbf{Scene}} &
  \multirow{2}{*}{\textbf{$\vect{\xi}$}} &
  \multirow{2}{*}{\textbf{Reg.}} &
  \multicolumn{1}{c|}{\multirow{2}{*}{\textbf{\# iter.}}} &
  \multicolumn{2}{c|}{\textbf{Depth error [mm]}} &
  \multicolumn{2}{c|}{\textbf{Depth error [\%]}} &
  \multicolumn{2}{c|}{\textbf{Normals error [deg]}} \\ \cline{5-10} 
 &
   &
   &
  \multicolumn{1}{c|}{} &
  \multicolumn{1}{c|}{\textbf{Mean}} &
  \multicolumn{1}{c|}{\textbf{Median}} &
  \multicolumn{1}{c|}{\textbf{Mean}} &
  \multicolumn{1}{c|}{\textbf{Median}} &
  \multicolumn{1}{c|}{\textbf{Mean}} &
  \multicolumn{1}{c|}{\textbf{Median}} \\ \hline
\multirow{2}{*}{01} &
  \multirow{2}{*}{$1/z$} &
  $\nabla$ &
  1 148 &
  \multicolumn{1}{r|}{1.0} &
  \textbf{\textless 0.1} &
  \multicolumn{1}{r|}{0.90} &
  \textbf{0.04} &
  \multicolumn{1}{r|}{1.19} &
  0.20 \\
 &
   &
  $\nabla^2$ &
  \textbf{73} &
  \multicolumn{1}{r|}{\textbf{0.3}} &
  \textbf{\textless 0.1} &
  \multicolumn{1}{r|}{\textbf{0.32}} &
  0.09 &
  \multicolumn{1}{r|}{\textbf{0.62}} &
  \textbf{0.18} \\ \hline
\multirow{2}{*}{02} &
  \multirow{2}{*}{$1/z$} &
  $\nabla$ &
  102 &
  \multicolumn{1}{r|}{0.2} &
  0.2 &
  \multicolumn{1}{r|}{0.35} &
  0.37 &
  \multicolumn{1}{r|}{1.00} &
  0.50 \\
 &
   &
  $\nabla^2$ &
  \textbf{64} &
  \multicolumn{1}{r|}{\textbf{0.1}} &
  \textbf{0.1} &
  \multicolumn{1}{r|}{\textbf{0.25}} &
  \textbf{0.21} &
  \multicolumn{1}{r|}{\textbf{0.95}} &
  \textbf{0.39} \\ \hline
\multirow{4}{*}{03} &
  \multirow{2}{*}{$1/z$} &
  $\nabla$ &
  \textgreater 5 550 &
  \multicolumn{1}{r|}{3.0} &
  2.1 &
  \multicolumn{1}{r|}{7.30} &
  6.73 &
  \multicolumn{1}{r|}{12.17} &
  9.44 \\
 &
   &
  $\nabla^2$ &
  \textgreater 1 500 &
  \multicolumn{1}{r|}{\textbf{1.9}} &
  \textbf{1.8} &
  \multicolumn{1}{r|}{5.83} &
  5.31 &
  \multicolumn{1}{r|}{\textbf{11.07}} &
  \textbf{8.00} \\
 &
  $d$ &
  $\nabla^2$ &
  301 &
  \multicolumn{1}{r|}{\textbf{1.9}} &
  \textbf{1.8} &
  \multicolumn{1}{r|}{5.85} &
  \textbf{4.99} &
  \multicolumn{1}{r|}{12.92} &
  9.15 \\
 &
  $1/d$ &
  $\nabla^2$ &
  \textbf{78} &
  \multicolumn{1}{r|}{\textbf{1.9}} &
  \textbf{1.8} &
  \multicolumn{1}{r|}{\textbf{5.78}} &
  5.21 &
  \multicolumn{1}{r|}{11.55} &
  8.30 \\ \hline
\end{tabular}%
}
\end{table}

The first experiment consists of a simple simulation, based on our photometric model as in \eqref{eq:simple-model}. We simulate a rotated plane, a curved surface, and a tubular geometry (see \fig{simple-dataset}).

\tab{simple-dataset} presents the results of this experiment. We conclude that the regularizer of the second derivative is better for accuracy (see in \fig{simple-0to2-convergence} the best result for each scene) and is also faster in convergence. 

Regarding the depth map variants, in a tubular geometry, inverse Euclidean distance $\vect{\xi}_{1/d}$ performs better than the alternatives. The corresponding results in \tab{simple-dataset} show much faster convergence with similar accuracy.

\subsection{Simulated colon dataset}
\label{sec:unity-dataset}

\begin{figure}
    \centering
    \includegraphics[width=\linewidth]{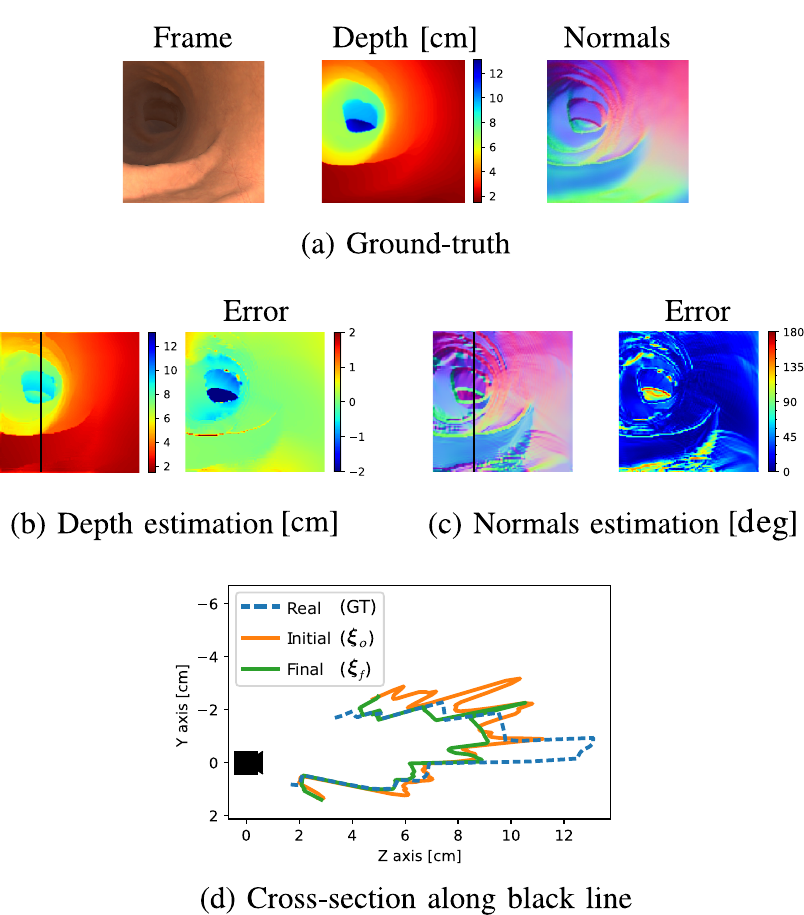}
    \caption{Results on the simulated colon dataset \cite{rau2019implicit}. Using variants $\vect{\xi}_{1/d}$ and $\nabla\vect{\xi}$. They show good accuracy for the estimated Z-depth map and the corresponding normals.}
    \label{fig:unity-dataset-results}
\end{figure}

\begin{table}
\caption{\textsc{Accuracy on simulated colon}~\cite{rau2019implicit}}
\label{tab:unity-dataset}
\centering
\resizebox{\linewidth}{!}{%
\begin{tabular}{|c|c|r|rr|rr|rr|}
\hline
\multirow{2}{*}{\textbf{$\vect{\xi}$}} &
  \multirow{2}{*}{\textbf{Reg.}} &
  \multicolumn{1}{c|}{\multirow{2}{*}{\textbf{\# iter.}}} &
  \multicolumn{2}{c|}{\textbf{Depth error [mm]}} &
  \multicolumn{2}{c|}{\textbf{Depth error [\%]}} &
  \multicolumn{2}{c|}{\textbf{Normals error [deg]}} \\ \cline{4-9} 
 &
   &
  \multicolumn{1}{c|}{} &
  \multicolumn{1}{c|}{\textbf{Mean}} &
  \multicolumn{1}{c|}{\textbf{Median}} &
  \multicolumn{1}{c|}{\textbf{Mean}} &
  \multicolumn{1}{c|}{\textbf{Median}} &
  \multicolumn{1}{c|}{\textbf{Mean}} &
  \multicolumn{1}{c|}{\textbf{Median}} \\ \hline
\multirow{2}{*}{$1/z$} &
  $\nabla$ &
  44 500 &
  \multicolumn{1}{r|}{5.3} &
  2.3 &
  \multicolumn{1}{r|}{10.60} &
  7.41 &
  \multicolumn{1}{r|}{30.63} &
  23.77 \\
 &
  $\nabla^2$ &
  8 900 &
  \multicolumn{1}{r|}{5.1} &
  4.0 &
  \multicolumn{1}{r|}{15.09} &
  11.86 &
  \multicolumn{1}{r|}{36.06} &
  29.26 \\ \hline
\multirow{2}{*}{$d$} &
  $\nabla$ &
  20 000 &
  \multicolumn{1}{r|}{3.3} &
  \textbf{1.6} &
  \multicolumn{1}{r|}{7.90} &
  \textbf{4.98} &
  \multicolumn{1}{r|}{\textbf{26.21}} &
  \textbf{18.75} \\
 &
  $\nabla^2$ &
  44 500 &
  \multicolumn{1}{r|}{3.8} &
  2.0 &
  \multicolumn{1}{r|}{9.57} &
  6.37 &
  \multicolumn{1}{r|}{32.00} &
  23.56 \\ \hline
\multirow{2}{*}{$1/d$} &
  $\nabla$ &
  44 500 &
  \multicolumn{1}{r|}{\textbf{2.8}} &
  \textbf{1.6} &
  \multicolumn{1}{r|}{\textbf{7.32}} &
  5.01 &
  \multicolumn{1}{r|}{27.89} &
  19.69 \\
 &
  $\nabla^2$ &
  \textbf{5 400} &
  \multicolumn{1}{r|}{5.1} &
  4.0 &
  \multicolumn{1}{r|}{15.16} &
  12.05 &
  \multicolumn{1}{r|}{35.86} &
  28.89 \\ \hline
\end{tabular}%
}
\end{table}

In this experiment, we validate our method on a frame of a photo-realistic dataset \cite{rau2019implicit}. This dataset simulates an endoscopy procedure based on a real CT scan of a human colon. The simulation includes effects more similar to those found in a real environment, such as richer textures and ambient light caused by secondary reflections, that are not considered in our model.

This input frame comes from an endoscope without distortion or vignetting and in which the light spread is homogeneous (see \fig{unity-dataset-results}a). We also know the average albedo of the surface and the gain of the endoscope. The results of this experiment are shown in the \tab{unity-dataset}. 

As in the previous case, we see that $\vect{\xi}_{1/d}$ is the best variant of the method, providing a good estimation (see \fig{unity-dataset-results}b) with less than 3~mm error on average. However, on this photo-realistic simulation, the $\nabla^2$ regularizer obtains lower accuracy. The second derivative is less robust to noise, such as that introduced by surface texture, which causes the albedo to be not perfectly uniform.

In addition, the photo-realistic simulator introduced a fog effect in areas far away from the camera. This increases the intensity of distant pixels. As a result, we cannot reconstruct the deepest part of the colon, from 10 to 12 cm (see \fig{unity-dataset-results}d). Therefore, the median error of 1.6~mm is considerably lower than the mean, which is influenced by those spurious.

\subsection{Real colon dataset}
\label{sec:hculb-dataset}

\begin{figure}
    \centering
    \setlength\tabcolsep{2pt}
    \begin{tabular}{cccc}
        \includegraphics[width=0.22\linewidth]{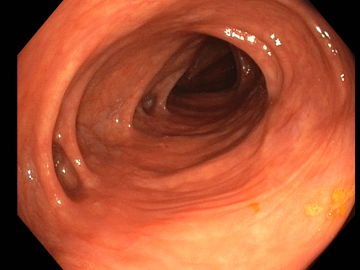} &
        \includegraphics[width=0.22\linewidth]{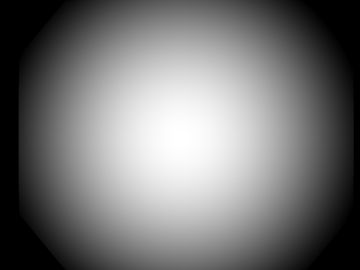} &
        \includegraphics[width=0.22\linewidth]{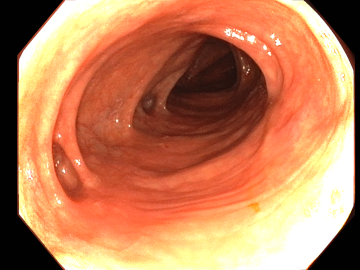} &
        \includegraphics[width=0.205\linewidth]{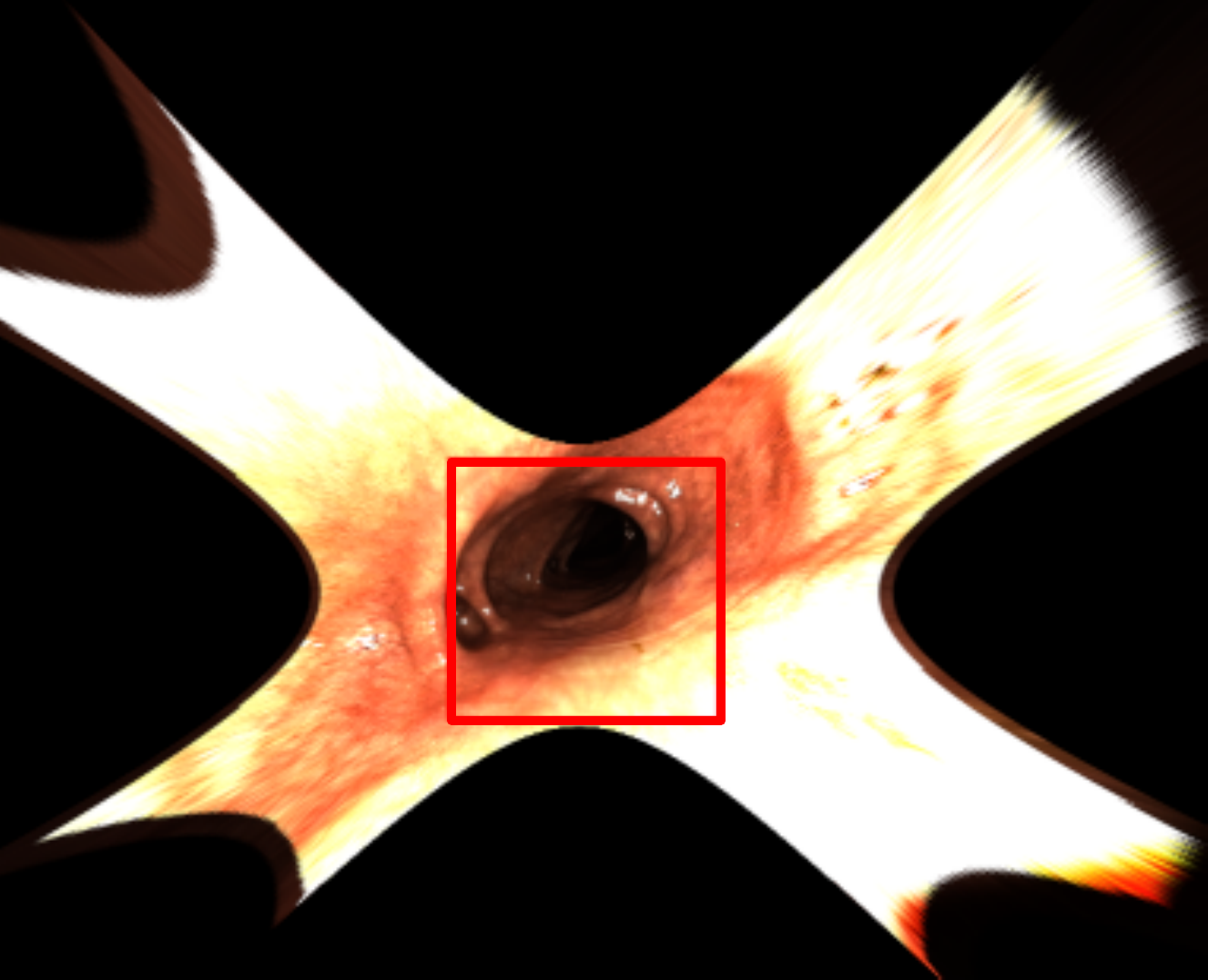} \\
        Original & Calibration & Canonical & Undistorted
    \end{tabular}
    \par\bigskip
    \setlength\tabcolsep{7pt}
    \begin{tabular}{ccc}
        \includegraphics[width=0.18\linewidth]{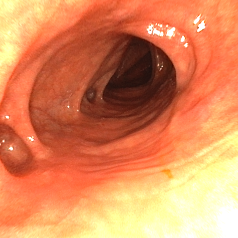} &
        \includegraphics[width=0.18\linewidth]{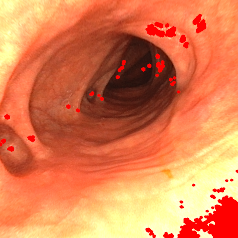} &
        \includegraphics[width=0.18\linewidth]{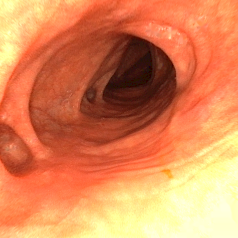} \\
        Cropped & Mask & Inpainted
    \end{tabular}
    \caption{HCULB real colon dataset from the EndoMapper project. We apply our photometric and geometric calibration and we perform highlight inpainting.}
    \label{fig:hculb-dataset}
\end{figure}

Our method is finally validated on a real image from the HCULB colonoscopy dataset (EndoMapper EU-H2020 project). The images of this dataset correspond to real human colons and are acquired \emph{in-vivo} during medical procedures, with the endoscope we calibrated in \sect{calibration}.

We take a single image (see \fig{hculb-dataset}). First, we compensate for calibrated vignetting and light spread function and obtain a frame with canonical illumination. Then, we undistort and crop the image. Finally, we perform automatic highlight detection and inpainting. With this, we obtain a frame similar to the one of the previous simulated dataset.

\fig{hculb-results} shows the reconstruction provided by our method. The estimated scale is arbitrary, as we do not have data about the automatic gain. Moreover, the HCULB dataset does not provide ground-truth information for comparison. Nevertheless, qualitatively, the result we have obtained properly reconstructs the tubular topology of the colon and also recovers notably the shape of the haustra.

\section{CONCLUSIONS}

This paper proposes a photometric stereo method that is able to reconstruct for the first time the geometry of the human colon using only the illumination on real monocular endoscopy procedures. We can recover the true scale of the environment if the surface albedo and the endoscope's auto-gain are known. The latter is set by the manufacturer of the hardware and the albedo could be reasonably estimated in the future since it is mostly uniform along the colon.

Our method obtains reconstructions with a mean accuracy below 3~mm on simulated data and is able to reconstruct the tubular geometry on a real colon, where it preserves the discontinuities at the colon's haustra.

In addition, a calibration process is designed to suit a medical endoscope. Our experiments show that we are able to model a real endoscope with an error of 3 gray levels. This allows us to conclude that our model, based on a virtual light source, offers a good compromise between accuracy and ease of calibration in a real environment.

Currently, depth estimation works in an off-line mode, but it allows us to overcome the lack of 3D perception inherent in monocular camera systems. In this way, for example, our method could constitute a new source of self-supervision for learning depth estimation without the need for stereo.

In conclusion, these results leave the door open to future work in real-time SLAM and autonomous navigation inside the colon, solving scale drift and allowing true scale maps.

\section*{ACKNOWLEDGMENT}

We acknowledge the work of Pablo Azagra (Universidad de Zaragoza) in performing the calibration sequence. We also thank Anita Rau (University College London) for providing preliminary data from her simulated dataset \cite{rau2019implicit}.

\bibliographystyle{IEEEtran} 
\bibliography{references}

\end{document}